\documentclass{bmvc2k}

\usepackage{graphicx}
\usepackage{pifont}
\usepackage{multicol}
\usepackage{multirow}
\usepackage{amsmath} 
\usepackage{newtxmath}
\usepackage{makecell}
\usepackage{booktabs}
\usepackage{xcolor}

\title{IRFusionFormer: Enhancing Pavement Crack Segmentation with RGB-T Fusion and Topological-Based Loss}

\addauthor{Ruiqiang Xiao}{rxiaoad@connect.ust.hk}{1}
\addauthor{Xiaohu Chen}{xiaohu@seu.edu.cn}{2}

\addinstitution{
 Robotics and Autonomous Systems\\
 Hong Kong University of Science and Technology (Guangzhou)\\
 Guangzhou, China
}
\addinstitution{
 Intelligent Transportation System Research Center\\
 Southeast University\\
 Nanjing, China\\
}

\runninghead{Xiao, Chen}{IRFusionFormer}

\begin{document}

\maketitle

\begin{abstract}
Crack segmentation is a critical task in civil engineering applications, particularly for assessing pavement integrity and ensuring the durability of transportation infrastructure. While deep learning models have advanced RGB-based segmentation, their performance degrades under adverse conditions like low illumination and motion blur. Thermal imaging offers complementary information by capturing emitted radiation, enabling better differentiation of cracks in challenging environments. By integrating information from both RGB and thermal images, RGB-T pavement crack segmentation has demonstrated significant advantages in complex real-world environments such as adverse weather conditions. However, research in this area remains relatively limited, and current RGB-T crack segmentation methods do not fully and efficiently leverage the complementary relationships between different modalities during multi-level information interaction.
To address this problem, we propose IRFusionFormer, a novel model for crack segmentation that effectively integrates RGB and thermal data. We introduce the Efficient RGB-T Cross Fusion Module (EGTCF) to capture extensive multi-scale relationships and long-range dependencies between modalities without incurring high computational costs. Additionally, we develop the Interaction-Hybrid-Branch-Supervision (IHBS) framework, which enhances modality interaction by distributing fused features across branches and enabling joint supervision. To preserve the topological structure of cracks, we propose a novel topology-based loss function that maintains connectivity and structural integrity during training.
Our method achieves state-of-the-art results, surpassing existing approaches with a Dice score of 90.01\% and an Intersection over Union (IoU) of 81.83\%. These advancements address critical challenges in pavement crack segmentation by improving robustness and accuracy under varying environmental conditions.
For access to the codes, data, and models pertinent to this study, please visit: \textit{\href{https://github.com/sheauhuu/IRFusionFormer}{Code}}.

\end{abstract}

\section{Introduction}
\label{sec:intro}
Crack segmentation, which involves assigning binary labels—crack or background—to individual pixels in an image, has attracted growing attention in various civil engineering scenarios such as buildings\cite{dais2021automatic}, bridges\cite{tran2023advanced}, tunnels\cite{wang2021pixel}, and pavement\cite{han2024crackdiffusion} inspections. Among these, pavement crack segmentation is particularly crucial for assessing road quality and maintaining the longevity. Traditional image processing techniques like thresholding\cite{li2008novel} and edge detection\cite{lokeshwor2014robust} have been used to segment pavement cracks in RGB images by exploiting grayscale value differences. However, these methods often suffer from low accuracy and lack robustness due to their reliance on handcrafted settings and sensitivity to varying imaging conditions.
The advent of deep learning, particularly convolutional neural networks (CNNs), has revolutionized semantic segmentation tasks. Methods like FCN\cite{long2015fully}, U-Net\cite{unet}, UNet++\cite{unetplusplus}, and DeepLab V3+\cite{deeplab} have achieved impressive results on large-scale RGB image semantic segmentation task. Consequently, researchers have adapted these paradigms for pavement crack segmentation. For instance, DeepCrack\cite{deepcrack} integrates multi-scale convolutional features from hierarchical stages to capture fine-grained line structures, leading to improved crack detection.

\begin{figure}[!ht]
  \centering
  \includegraphics[width=0.8\textwidth]{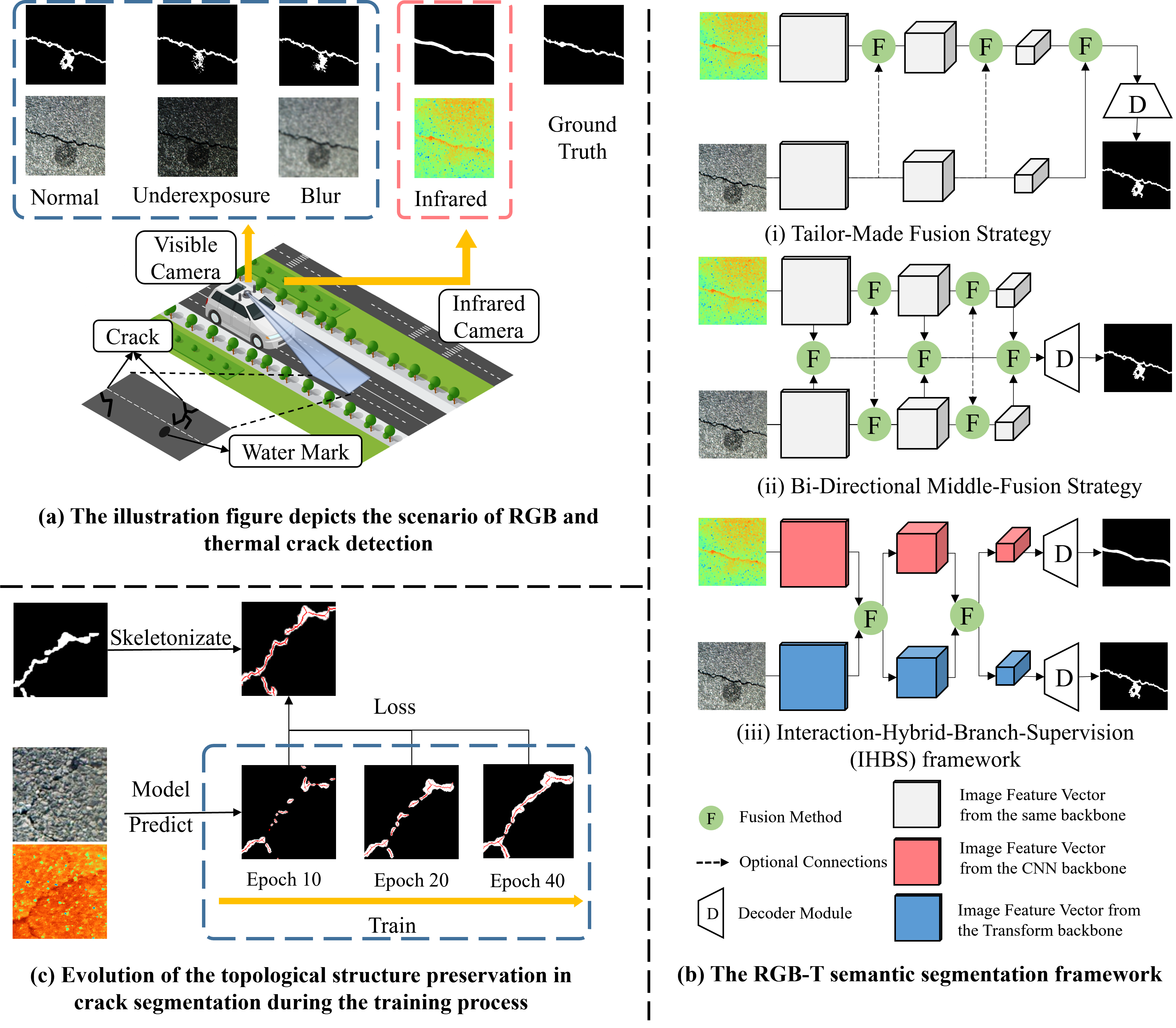}
  \caption{The illustration figure comprises (a) The illustration figure depicts the scenario of RGB and thermal crack detection, (b) The RGB-T semantic segmentation framework and (c) Evolution of the topological structure preservation in crack segmentation during the training process.}\label{fig:intro}
\end{figure}

Despite these advancements, RGB-based pavement crack segmentation methods degrade rapidly under challenging conditions such as rainy or hazy weather and low illumination\cite{fan2023pavement}. During pavement inspections, cameras mounted on fast-moving vehicles like inspection trucks or drones struggle to maintain stable footage, minimize motion blur, and ensure adequate illumination simultaneously. Consequently, underexposed and blurred RGB images yield unfavorable segmentation results compared to normal conditions, as shown in the \textbf{Figure \ref{fig:intro} (a)}. Additionally, semantic interferences resembling cracks—such as scattered binding particles, tree shadows, water marks, and patch repairs—introduce additional complexities, leading to false detections and reduced reliability\cite{wangComplexScenePavement2023}.
In contrast, thermal images rely on emitted radiation from objects and can capture stable images under complex conditions, albeit at lower resolutions. This capability allows for better differentiation between crack foregrounds and backgrounds in challenging environments\cite{du2017detection, jiang2024asphalt}. Therefore, RGB-Thermal (RGB-T) crack segmentation has gained increased attention\cite{Kutuk_2022_CVPR}. By utilizing both RGB and thermal data and efficiently fusing their complementary information, the performance and stability of crack segmentation can be enhanced across various real-world scenarios. Leveraging the rich semantic information from both modalities is essential to overcome existing challenges\cite{tang2022image}. Although some studies have employed both infrared and RGB images, a unified benchmark for pavement crack segmentation across these modalities is still lacking. To address this gap, we compiled existing crack segmentation approaches and hybrid methods integrating infrared and RGB data, constructing a novel and comprehensive benchmark for asphalt pavement crack segmentation. Our review identified several critical limitations in current methodologies.

One of the key challenges in RGB-T crack segmentation is the efficient fusion of cross-modal features. Traditional convolution-based attention mechanisms—such as channel, spatial, and hybrid attention—are limited by their local receptive fields, restricting their ability to model global contextual relationships\cite{chen2024globalsr}. While self-attention mechanisms can capture long-range dependencies, their high computational and memory costs make them impractical for large-scale inputs\cite{efficient}. To address these issues, we introduce the Efficient RGB-T Cross Fusion Module (EGTCF), which effectively captures extensive multi-scale relationships and long-range feature interdependencies between RGB and thermal modalities without incurring prohibitive computational overhead.
Another challenge lies in designing an effective framework for multi-modal feature learning and supervision. Existing RGB-T segmentation frameworks can be divided into two main types(illustrated in \textbf{Figure \ref{fig:intro}(b)}): the tailor-made under-fused strategy and the bi-directional middle-fusion strategy\cite{wu2022complementarity}. The tailor-made approach integrates thermal features into RGB features within the encoder but lacks sufficient inter-modal interaction. The bi-directional middle-fusion strategy promotes interaction by allowing fused features to influence unimodal branches but struggles with supervising modality-specific learning effectively. To overcome these limitations, we propose the Interaction-Hybrid-Branch-Supervision (IHBS) framework. This framework enhances modality interaction by distributing fused feature information across branches and enabling joint supervision of RGB and thermal feature learning.
Furthermore, an appropriate loss function is also critical for enhancing deep learning performance in crack segmentation models.\cite{wang2020comprehensive}. Commonly used loss functions in crack segmentation, such as Cross Entropy Loss and Dice Loss\cite{jadon2020survey}, focus on pixel-level prediction accuracy but often neglect the topological structure of cracks\cite{wang2018pavement}, leading to discontinuities in the segmented outputs. To remedy this, a topological-based loss function, illustrated in \textbf{Figure \ref{fig:intro}(c)}, is introduced to accurately capture and preserve the crack skeleton's intrinsic topology.  

In summary, the key contributions of our research are outlined as follows:
\begin{itemize}
    \item We propose a novel method called IRFusionFormer for crack segmentation, achieving state-of-the-art results on benchmark datasets. Specifically, our framework outperforms existing approaches by attaining Dice and Intersection over Union (IoU) scores of 90.01\% and 81.83\%, respectively. This demonstrates its effectiveness in accurately identifying and delineating cracks, thereby enhancing both the efficiency and reliability of pavement maintenance practices.
    \item We propose the deployment of the Efficient RGB-T Cross Fusion Module (EGTCF) to capture multi-scale extensive relationships and ong-range feature interdependencies between RGB and thermal modalities. Additionally, the Interaction-Hybrid-Branch-Supervision (IHBS) framework facilitates the sharing of fused feature information across multiple branches and supports simultaneous supervision of feature learning in different modalities.
    \item We introduce a topology-based loss function aimed at preserving the connectivity and topological structure of asphalt pavement cracks. This innovation significantly advances the accuracy and consistency of crack segmentation by systematically integrating topological considerations into the learning process.
\end{itemize}

These advancements significantly contribute to the field of pavement maintenance by improving the accuracy and efficiency of crack detection technologies, which are vital for prolonging pavement lifespan and ensuring road safety.

\section{Methods}

\subsection{Overview} \label{overview}
\begin{figure}[!ht]
  \centering
  \includegraphics[width=0.8\textwidth]{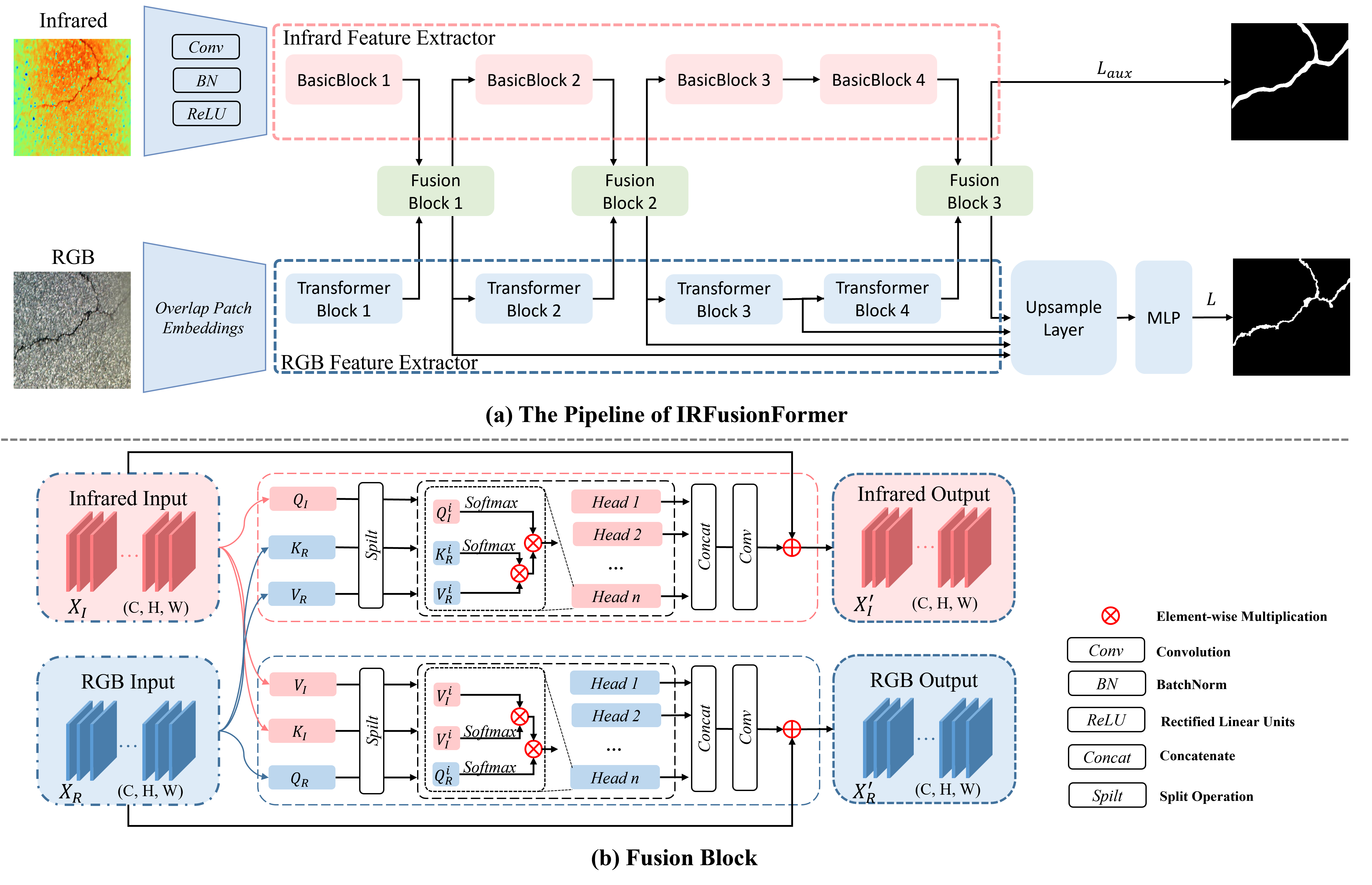}
  \caption{The IRFusionFormer framework comprises (a) the pipeline of the framework and (b) fusion block using efficient cross attention mechanism}\label{fig:framework}
\end{figure}

In modern image feature extraction, CNN and Transformer-based architectures serve as the primary methods.
The CNN architecture is effective in extracting local features, but due to its limited receptive field and the pooling process, it may miss some global-scale correlations. 
In contrast, the Transformer architecture uses a self-attention mechanism to capture long-range dependencies and global context, enabling a comprehensive understanding of semantic entities. 
The proposed IHBS framework requires separate feature extraction for infrared and RGB images, followed by an interaction mechanism that enhances modality fusion and enables joint supervision.
Infrared images, sensitive to thermal characteristics, highlight areas with temperature differences, while RGB images provide detailed visual information. To achieve this, as shown in \textbf{Figure~\ref{fig:framework}}, ResNet is used for infrared image feature extraction, capturing local crack information, while the Segformer network, based on Transformer technology, is employed for RGB images, capturing both local and global features. After processing through the 1st, 2nd, and 4th ResNet or Segformer decoders, the infrared and RGB images of identical size are fused via the EGTCF.The fused features are then redistributed back into their respective modality-specific branches, enhancing inter-modal interaction as part of the IHBS framework, thereby promoting a more effective modality interaction.
In the training phase, the IHBS framework employs joint supervision of RGB and thermal feature learning, while both the topological-based loss function and the infrared-based auxiliary loss function guide the segmentation training.

\subsection{Efficient RGB-T Cross Fusion Module} \label{fusion}

Incorporating both thermal and RGB information has been shown to enhance segmentation performance\cite{sun2020fuseseg}. Current techniques utilizing convolution or pooling layers have led to restricted receptive fields, limiting the exploration of relationships between RGB images and thermal maps. Moreover, the self-attention mechanism is unsuitable for processing extensive inputs like shallow RGB and infrared feature maps. Instead of employing a basic non-local fusion strategy, we have devised the Enhanced Global Thermal and Color Feature (EGTCF) module for integrating multimodal features, as illustrated in \textbf{Figure \ref{fig:framework}}.

For infrared feature maps denoted as $X_{I} \in \mathbb{R}^{C\times H \times W}$ and RGB features denoted as $X_{R} \in \mathbb{R}^{C\times H \times W}$, we initially pass both features through separate convolutional layers to generate query tensor, key tensor, and value tensor. These are denoted as $Q_{I} \in \mathbb{R}^{C_{k} \times H \times W}$, $K_{I} \in \mathbb{R}^{C_{k} \times H \times W}$, $V_{I} \in \mathbb{R}^{C_{v} \times H \times W}$ for infrared features and $Q_{R} \in \mathbb{R}^{C_{k} \times H \times W}$, $K_{R} \in \mathbb{R}^{C_{k} \times H \times W} $, $V_{R} \in \mathbb{R}^{C_{v} \times H \times W}$ for visible features, as shown in Equation \ref{eq:dot}. Here, $C_{k}$ and $C_{v}$ correspond  to the dimensions of the convolution matrix ($W_{k}$, $W_{q}$, $W_{v}$). 

\begin{equation}
\label{eq:dot}
\begin{aligned}
    Q_{I} &= W_{q}^{I} \cdot {X}_{I}, & K_{I} &= W_{k}^{I} \cdot {X}_{I}, & V_{I} &= W_{v}^{I} \cdot {X}_{I}, \\
    Q_{R} &= W_{q}^{R} \cdot {X}_{R}, & K_{R} &= W_{k}^{R} \cdot {X}_{R}, & V_{R} &= W_{v}^{R} \cdot {X}_{R}
\end{aligned}
\end{equation}

Drawing inspiration from the work of \cite{efficient}, a cross-efficient fusion module has been implemented to capture semantic relationships across multimodal features by leveraging an efficient attention mechanism that addresses long-range dependencies while minimizing memory and computational complexities. Initially, the query, key, and value tensors are partitioned into $n$ segments to streamline computational operations. For instance, in the case of the RGB query tensor, segments are defined as $Q_{R}^{i} = Q_{R}^{\left[ \dfrac{C_{k}}{n} \cdot i, \dfrac{C_{k}}{n} \cdot (i+1) \right]} \in \mathbb{R}^{\dfrac{C_{k}}{n} \times H \times W}$. This split approach is also applied to the query, key, and value tensors associated with the infrared and RGB features.
Subsequently, the spatial dimensions of $V^{i}$ and $K^{i}$ are flattened to $\hat{V^{i}} \in \mathbb{R}^{\dfrac{C_{v}}{n} \times HW}$ and $\hat{K^{i}} \in \mathbb{R}^{\dfrac{C_{k}}{n} \times HW}$. The cross attention matrix $A_{R}$ and $A_{I}$ are computed by multiplying $\hat{V^{i}}$ with the normalized $\hat{K^{i}}$ from the complementary modality through the softmax function: $A_{R}^{i} = \text{softmax}(\hat{K_{I}^{i}})^{T} \cdot \hat{V_{I}^{i}} \quad$ and $A_{I}^{i} = \text{softmax}(\hat{K_{R}^{i}})^{T} \cdot \hat{V_{R}^{i}}$.
Next, we multiply the attention matrix $A$ with the normalized query tensor $Q_{R}$ within the same modality, concatenating all channels, applying a projection convolution layer, and adding the input feature residuals to generate new fusion features from the efficient RGB-T cross-fusion module:
\begin{equation}
\label{eq:attention}
\begin{aligned}
    X_{I}^{'} &= \text{conv(concat(softmax}(Q_{I}^{i}) \cdot A_{I}^{i}) \|_{i \in \left [ 0, n-1\right ]}) + X_{I}, \\
    X_{R}^{'} &= \text{conv(concat(softmax}(Q_{R}^{i}) \cdot A_{R}^{i}) \|_{i \in \left [ 0, n-1\right ]}) + X_{R}
\end{aligned}
\end{equation}
This efficient RGB-T cross-fusion module facilitates feature interactions between parallel streams during each stage of the feature extraction process. It enables the learning of long-range dependencies from the other modality by correcting its own modal features through these interactions. Employing this methodology significantly reduces computational complexity. For instance, while a non-local module applied to a 256*256 image would require 17GB of memory, our efficient attention approach demands only 67 MB of memory.

\subsection{Topological-based Loss Function} \label{loss}
In the segmentation of tubular objects such as pavement cracks, considering their topological structure can significantly enhance the usability of the segmentation results\cite{shit2021cldice}.
Therefore, our research proposes the integration of a topological-based loss function into the segmentation process, aiming to preserve the structural integrity of cracks. 
In the context of asphalt pavement crack detection, the actual mask is referred to as $V_L$, while the model-generated predicted mask is symbolized as $V_P$.
Maximum pooling is applied to refine images by smoothing object boundaries and removing minor noise. 
The difference between the images before and after this operation highlights the skeleton features, which are further refined through multiple iterations.
The skeleton derived from the actual ground truth mask is $S_L$, while that from the model's prediction is termed as $S_P$.
Topological precision is defined as $T_{\text{precision}}=\frac{|S_P\cap V_L|}{|S_P|}$, which is significantly impacted by False Positives (FP). And topological sensitivity is defined as $T_{\text{sensitivity}}=\frac{|S_L\cap V_P|}{|S_L|}$, which is markedly affected by False Negatives (FN). The loss function $L_{\text{Topology}}$ is calculated using the harmonic mean of topological precision and sensitivity: 
\begin{equation*}
    L_{\text{Topology}}(V_L,V_P)=2\times\frac{T_{\text{precision}}(S_P,V_L) \times T_{\text{sensitivity}}(S_L,V_P)}{T_{\text{precision}}(S_P,V_L)+T_{\text{sensitivity}}(S_L,V_P)}
\end{equation*}
However, $L_{\text{Topology}}$ is primarily focused on the overall continuity and connectivity, which can result in challenges when attempting to accurately segment crack edges within images.
Consequently, in the proposed loss function, the topological-base loss function $L_{\text{Topology}}$ is combined with Cross Entropy Loss $L_{\text{CE}}$ and Dice Loss $L_{\text{Dice}}$ using the weights $\alpha$, $\beta$ and $\gamma$, thus constructing a more robust framework for crack segmentation.
This composite loss function leverages the strengths of each component: $L_{\text{Topology}}$ for maintaining topological integrity, $L_{CE}$ for pixel-wise accuracy, and $L_{\text{Dice}}$ for optimizing over imbalanced data conditions.
Additionally, in the training phase of the proposed framework, features extracted from infrared images are fused with RGB image features to produce a fusion segmentation result, which is denoted as $V_P^I$. Unlike the final output, this result is based one the feature of the infrared images, incorporating their inherent physical constraint features. To further capitalize on these infrared-specific constraints and enhance crack segmentation, the proposed loss function includes an auxiliary loss function, denoted as $L_{\text{aux}}(V_L,V_P^I)$. Ultimately, the proposed loss function consists of the aforementioned loss functions:
\begin{equation}
\label{eq:Loss}
\begin{aligned}
    \centering
    && &L(V_L,V_P)=\alpha L_{\text{Topology}}(V_L,V_P)+\beta L_{\text{CE}}(V_L,V_P)+\gamma L_{\text{Dice}}(V_L,V_P) &\\
    && &L_{\text{aux}}(V_L,V_P^I)=\alpha L_{\text{Topology}}(V_L,V_P^I)+\beta L_{\text{CE}}(V_L,V_P^I)+\gamma L_{\text{Dice}}(V_L,V_P^I), &\\
    && &L(V_L,V_P,V_P^I)=L(V_L,V_P) + \delta L_{\text{aux}}(V_L,V_P^I) &
\end{aligned}
\end{equation}
where $\alpha$, $\beta$, $\gamma$ and $\delta$ represent the weights for $L_{\text{Topology}}$, $L_{\text{CE}}$, $L_{\text{Dice}}$ and $L_{\text{aux}}$.

\section{Experiments}

\subsection{Dataset}

The dataset employed in this study is an open-source dataset\cite{liu2022asphaltpavementcrack} dedicated to crack detection using Infrared Thermography (IRT), which is included in the RGB-T asphalt pavement crack segmentation benchmark. It comprises four image types: RGB images, infrared images, fused images (combined at a 50:50 ratio using IR-Fusion™ technology), and ground truth images manually annotated using Photoshop. Each category consists of 448 images, each with a resolution of 640x480 pixels. For training and evaluation purposes, the segmentation model divides the entire dataset into two subsets: 358 images for the training set and 90 images for the test set.

\subsection{Training Details}
To enhance the diversity of the training data and improve the model's robustness, spatial, color, and numerical transformations were applied to the training set, such as random horizontal or vertical flips in spatial, randomly altering brightness or contrast in color. For images in the validation set, only resizing to 480×480 pixels and normalization processes were applied. The proposed IRFusionFormer was implemented using the PyTorch framework and optimized with AdamW, incorporating a weight decay of 1e-4. A batch size of 8 and 150 training epochs were designated for training. All experiments were performed on an NVIDIA GeForce RTX 4090 to expedite model training.

\subsection{Evaluation Metrics}

To accurately and objectively evaluate the performance of various models, we employed six widely used evaluation metrics in image segmentation: Dice, IoU (Intersection over Union), Accuracy, Precision, Specificity, and Recall. Higher values for these metrics indicate superior segmentation performance. Simultaneously, these metrics are also used as the evaluation criteria for the RGB-T asphalt pavement crack segmentation benchmark.

\section{Results}
\subsection{Comparison with State-of-the-Art Methods}

\begin{figure}[!ht]
  \centering
  \includegraphics[width=0.9\textwidth]{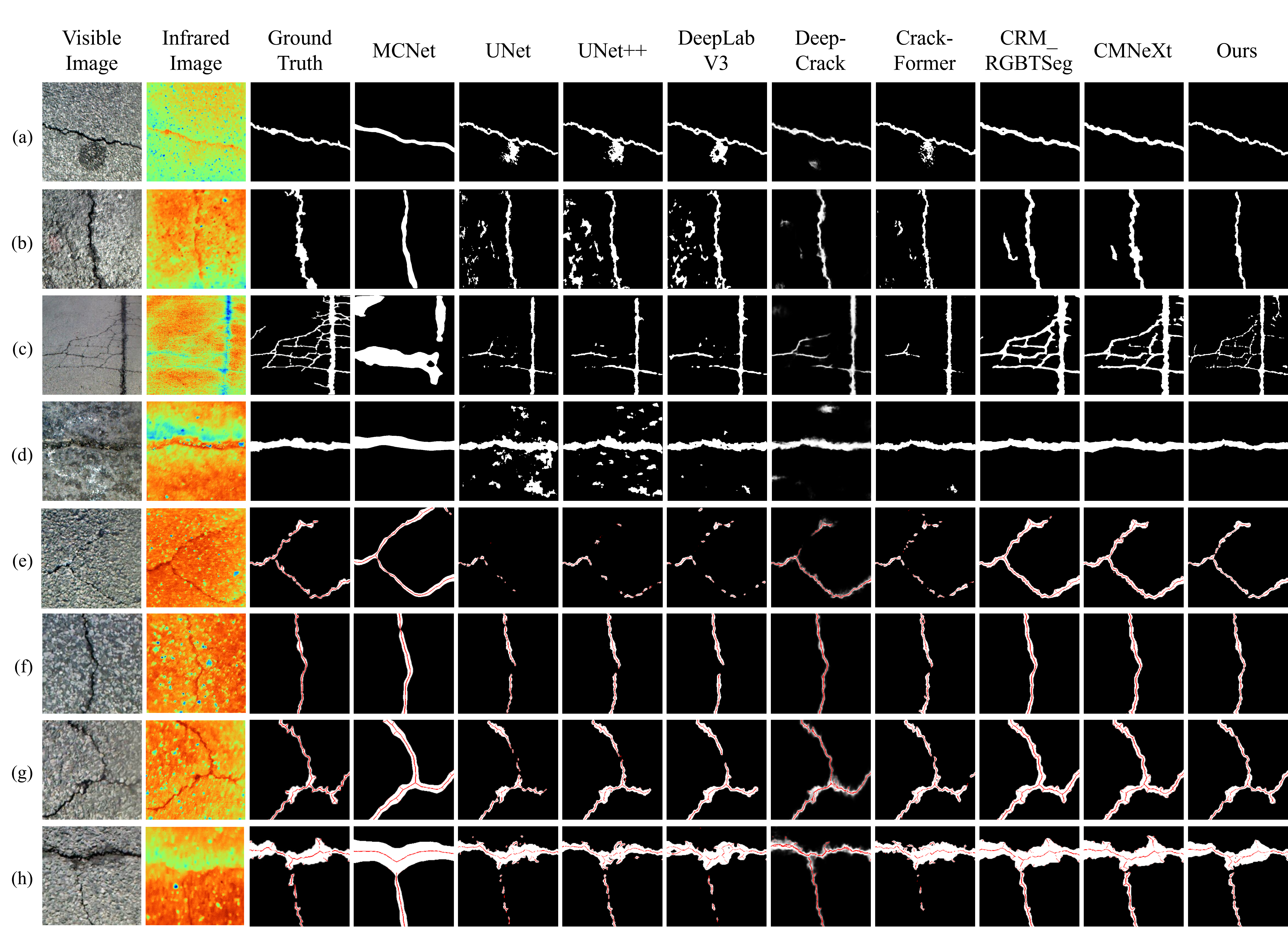}
  \caption{Comparison of crack segmentation results from nine models on test set of the dataset. Images (e)-(h) illustrate the skeleton of the cracks.}\label{fig:crack}
\end{figure}

To validate the effectiveness of the proposed method, we compared it against eight mainstream models on the dataset. Among these, MCNet\cite{mcnet} uses only infrared images as input, whereas U-net\cite{unet}, UNet++\cite{unetplusplus}, DeepLabV3\cite{deeplab}, DeepCrack\cite{deepcrack}, and CrackFormer\cite{CrackFormer} use solely RGB images. CRM\_RGBT\_Seg\cite{crm} and CMNeXt\cite{cmx} employ both infrared and RGB images as inputs. IRFusionFormer, along with eight comparative models, constitutes the benchmarking suite of the proposed benchmark. The quantitative results are summarized in \textbf{Table \ref{tab:main_results}} and demonstrate that our proposed IRFusionFormer outperforms other SOTA methods on the dataset.
\textbf{Figure ~\ref{fig:crack}} displays the visual comparsion of 9 models on the test sets of the datasets, with some results highlighting the skeleton of the cracks.
\begin{table}[!ht]\footnotesize
\caption{Quantitative results on various datasets. Best and second-best results are bold and underlined, respectively. For type, 'I' stands for Infrared, 'R' stands for RGB, and 'IR' stands for Infrared+RGB.}\label{tab:main_results}
\begin{center}
\begin{tabular}{l l l l l l l l}
\toprule
Type & Models(Venue) & Dice & IoU & Accuracy & Precision & Specificity & Recall \\\hline
I & \makecell{MCNet\cite{mcnet}} & 0.6844 & 0.5202 & 0.9467 & 0.7482 & 0.9786 & 0.6306 \\
R & \makecell{U-net\cite{unet}} & 0.7891 & 0.6517 & 0.9794 & 0.8161 & 0.9909 & 0.7639 \\
R & \makecell{UNet++\cite{unetplusplus}} & 0.8048 & 0.6733 & 0.9801 & 0.7937 & 0.9887 & 0.8574 \\
R & \makecell{DeepLabV3\cite{deeplab}} & 0.8338 & 0.7149 & 0.9828 & 0.8134 & 0.9896 & 0.8552 \\
R & \makecell{DeepCrack\cite{deepcrack}} & 0.7406 & 0.5880 & 0.9787 & 0.6592 & 0.9837 & 0.8450 \\
R & \makecell{CrackFormer\cite{CrackFormer}} & 0.8489 & 0.7374 & \underline{0.9847} & 0.8462 & 0.9918 & 0.8515 \\
IR & \makecell{CRM\_RGBTSeg\cite{crm}} & 0.8450 & 0.7370 & 0.9829 & 0.8651 & \underline{0.9921} & 0.8293 \\
IR & \makecell{CMNeXt\cite{cmx}} & \underline{0.8760} & \underline{0.7794} & 0.9835 & \underline{0.8885} & \underline{0.9921} & \underline{0.8639} \\
IR & \makecell{\textit{IRFusionFormer}\textit{(ours)}} & \textbf{0.9001} & \textbf{0.8183} & \textbf{0.9899} & \textbf{0.9001} & \textbf{0.9947} & \textbf{0.9001} \\
\bottomrule
\end{tabular}
\end{center}
\end{table}

Results in \textbf{Table \ref{tab:main_results}} indicate that the RGB-infrared integrated model outperforms the model that uses only RGB images, both of which are superior to the model that uses only infrared images. As seen in \textbf{Figure \ref{fig:crack}}, MCNet's segmentation results are smoother at the edges compared to other networks, indicating a lack of sufficient information. The proposed IRFusionFormer achieved the best results across all six evaluation metrics on the test dataset, outperforming the second-best model, CMNeXt, in Dice, IoU, Accuracy, Precision, Specificity, and Recall by 2.41\%, 3.89\%, 0.64\%, 1.16\%, 0.26\%, and 3.62\%, respectively. More importantly, infrared-integrated models were less affected by the presence of watermarks, shadows, or other disturbances on the pavement, compared to RGB-only models. These findings suggest that while infrared images alone are insufficient in isolation for accurate asphalt pavement crack segmentation, their integration with RGB images can markedly enhance prediction accuracy and reduce the impact of complex environmental factors. Additionally, as illustrated in \textbf{Figure ~\ref{fig:crack}}, the proposed method provides a more complete crack skeleton, attributed to the comprehensive crack information from infrared images and the specially designed loss function that accounts for the topology of cracks and integrates auxiliary loss from infrared feature fusion.

\subsection{Ablation Study}

Our IRFusionFormer mainly consists of two parts:  the Efficient RGB-T Cross Fusion Module and the Topological-based Loss Function.  Therefore, we conduct ablation studies to verify the effectiveness of each component, and then analyze the stages of fusion modules and loss function weights. Each of the three fusion modules in the model was evaluated both individually and in combination, with results detailed in \textbf{Table \ref{tab:fusion_results}}. Additionally, the impact of varying weights of the topological-based loss function and the incorporation of auxiliary loss functions was examined, with findings presented in \textbf{Table \ref{tab:aux_loss_results}} and \textbf{Table \ref{tab:topology_loss_results}}.

\begin{table}[!ht]\footnotesize
    \caption{Quantitative results on various fusion stages. Best and second-best results are bold and underlined, respectively.}\label{tab:fusion_results}
    \begin{center}
    \begin{tabular}{ccccccccc}
        \toprule
            \multicolumn{3}{c}{Fusion}  & \multirow{2}{*}{Dice} & \multirow{2}{*}{IoU} & \multirow{2}{*}{Accuracy} & \multirow{2}{*}{Precision} & \multirow{2}{*}{Specificity} & \multirow{2}{*}{Recall} \\ \cline{1-3}
            Stage 0     & Stage 1   & Stage 2   &               &                      &                           &                           &                              &                         \\ \hline
            \ding{51}   &           &           & 0.8706 & 0.7709 & 0.9869 & 0.8644 & 0.9927 & 0.8770 \\
                        & \ding{51} &           & 0.8717 & 0.7726 & 0.9870 & 0.8680 & 0.9929 & 0.8755 \\
                        &           & \ding{51} & 0.8737 & 0.7757 & 0.9872 & 0.8680 & 0.9929 & 0.8794 \\
            \ding{51}   & \ding{51} &           & 0.8819 & 0.7887 & 0.9881 & 0.8817 & 0.9937 & 0.8821 \\
            \ding{51}   &           & \ding{51} & 0.8838 & 0.7917 & 0.9883 & 0.8852 & 0.9939 & \underline{0.8823} \\
                        & \ding{51} & \ding{51} & \underline{0.8841}& \underline{0.7923} & \underline{0.9884} & \underline{0.8863} & \underline{0.9940} & 0.8820 \\
            \ding{51}   & \ding{51} & \ding{51} & \textbf{0.9001}   & \textbf{0.8183} & \textbf{0.9899} & \textbf{0.9001} & \textbf{0.9947} & \textbf{0.9001} \\
        \bottomrule
\end{tabular}
\end{center}
\end{table}
Analysis of \textbf{Table \ref{tab:fusion_results}} reveals that experimental outcomes improve as the number of Fusion Blocks increases within the feature extraction network. Using Fusion Blocks across all three stages yielded optimal results. Notably, the application of a Fusion Block at the third stage, corresponding to the high-level feature stage, more effectively captures information from the alternate modality than at the low-level stages. 

\begin{table}[!ht]\footnotesize
    \caption{Quantitative results with varying auxiliary loss function weight. Best and second-best results are bold and underlined, respectively.}\label{tab:aux_loss_results}
    \begin{center}
        \begin{tabular}{cccccccc}
            \toprule
            \textbf{Aux Loss} & \textbf{Dice} & \textbf{IoU} & \textbf{Accuracy} & \textbf{Precision} & \textbf{Specificity} & \textbf{Recall} \\
            \midrule
            0   & 0.8759 & 0.7792 & 0.9875 & 0.8752 & 0.9934 & 0.8766 \\
            0.1 & \textbf{0.9001}   & \textbf{0.8183} & \textbf{0.9899} & \underline{0.9001} & \underline{0.9947} & \textbf{0.9001} \\
            0.2 & \underline{0.8886} & \underline{0.7996} & \underline{0.9889} & 0.8949 & 0.9945 & 0.8824 \\
            0.3 & 0.8875 & 0.7978 & 0.9888 & \textbf{0.9005} & \textbf{0.9949} & 0.8748 \\
            0.4 & 0.8857 & 0.7948 & 0.9885 & 0.8898 & 0.9942 & 0.8816 \\
            0.5 & 0.8834 & 0.7912 & 0.9883 & 0.8836 & 0.9938 & \underline{0.8833} \\
            \bottomrule
        \end{tabular}
    \end{center}
\end{table}

\begin{table}[!ht]\footnotesize
    \caption{Quantitative results with varying and topological-based loss function weight. Best and second-best results are bold and underlined, respectively.}\label{tab:topology_loss_results}
    \begin{center}
        \begin{tabular}{cccccccc}
            \toprule
            \textbf{$L_{\text{topology}}$ Weight} & \textbf{Dice} & \textbf{IoU} & \textbf{Accuracy} & \textbf{Precision} & \textbf{Specificity} & \textbf{Recall} \\
            \midrule
            0   & 0.8792 & 0.7844 & 0.9881 & \underline{0.8989} & \textbf{0.9949} & 0.8604 \\
            0.1 & 0.8842 & 0.7924 & 0.9885 & 0.8959 & 0.9946 & 0.8728 \\
            0.2 & \underline{0.8886} & \underline{0.7996} & \underline{0.9889} & 0.8949 & 0.9945 & 0.8824 \\
            0.3 & \textbf{0.9001}  & \textbf{0.8183} & \textbf{0.9899} & \textbf{0.9001} & \underline{0.9947} & \textbf{0.9001} \\
            0.4 & 0.8861 & 0.7955 & 0.9884 & 0.8766 & 0.9933 & \underline{0.8958} \\
            0.5 & 0.8815 & 0.7881 & 0.9880 & 0.8793 & 0.9936 & 0.8836 \\
            0.6 & 0.8729 & 0.7745 & 0.9871 & 0.8682 & 0.9929 & 0.8776 \\
            \bottomrule
        \end{tabular}
    \end{center}
\end{table}

From \textbf{Table \ref{tab:aux_loss_results}}, it is evident that the use of the auxiliary loss function significantly improves segmentation performance. However, when the weight of the auxiliary loss function increases, the results of crack segmentation decline. Specifically, when the weight is 0.1, the results are optimal.
From \textbf{Table \ref{tab:topology_loss_results}}, the topological-based loss function demonstrates an enhanced effect on crack segmentation. However, excessively high weights of the topological-based loss function lead to decreased segmentation performance. Consequently, an appropriately weighted topological-based loss function optimally enhances the model’s segmentation performance. The ablation study shows, a weight of 0.3 yields the best performance. Overall, these results demonstrate that the proposed multi-scale fusion module and appropriately weighted loss functions significantly contribute to the performance of the crack detection model.

\section{Conclusions}

To integrate RGB and infrared image information for segmenting asphalt pavement cracks in complex environments, we propose the RGB-T asphalt pavement crack segmentation benchmark. The benchmark includes a dataset of image pairs, a codebase comprising nine algorithms, six evaluation metrics, as well as all related results. This benchmark provided a new platform for various methods. IRFusionFormer, which was a new crack segmentation method proposed in this research, achieved state-of-the-art (SOTA) results in the the established benchmark. In the proposed method, the Efficient RGB-T Cross Fusion Module was incorporated into the Interaction-Hybrid-Branch-Supervision (IHBS) framework, which was designed to efficiently fuse features from RGB and infrared images across three key stages. Additionally, a topological-based loss function was employed, specifically tailored to handle the topological structures of cracks, thereby improving the accuracy and robustness of the crack segmentation. Ablation study results demonstrated that these techniques significantly improve the performance of the IRFusionFormer network in crack segmentation tasks.

\bibliography{bmvc_final}

\end{document}